%% file: main.tex
\documentclass{article}

\PassOptionsToPackage{numbers}{natbib}

\usepackage[preprint]{neurips_2024}

\usepackage{subfiles}

\usepackage{mwe}
\usepackage[utf8]{inputenc} 
\usepackage[T1]{fontenc}    
\usepackage{hyperref}       
\usepackage{csquotes}
\usepackage{url}            
\usepackage{booktabs}       
\usepackage{amsfonts}       
\usepackage{nicefrac}       
\usepackage{microtype}      
\usepackage{xcolor}         
\usepackage{amsthm}
\usepackage{caption}
\usepackage{amssymb}
\usepackage{makecell}
\usepackage{varwidth}

\usepackage{graphicx}
\usepackage{subfigure}

\usepackage{parskip}
\usepackage{tikz}
\usepackage{lipsum}
\usepackage{bbm}
\usepackage{algorithm}
\usepackage{algorithmic}
\usepackage{amsmath}
\usepackage{multirow}
\usepackage{hhline}
\usepackage{hyperref}
\usepackage{mathtools}
\usepackage[compact]{titlesec}

\theoremstyle{definition}

\newcommand{\E}{\mathbb{E}}

\newcommand{\vecX}{\textbf{x}}
\newcommand{\Xx}{\textbf{$\mathcal{X}$}}
\newcommand{\Y}{\textbf{$\mathcal{Y}$}}
\newcommand{\T}{\textbf{$\mathcal{T}$}}

\newcommand{\D}{\textbf{\textit{D}}}

\DeclareMathOperator*{\argmin}{arg\,min}

\title{Leveraging Data from Disparate Sources for ITE Estimation in Federated Learning Settings}
\author{
  Disha Makhija \\
  Electrical and Computer Engineering\\
  University of Texas at Austin\\
  Austin, TX 78705 \\
  \texttt{disham@utexas.edu} \\
    \And
   Yejin Kim \\
   University of Texas Health at Houston \\
   Houston, TX 78705 \\
   \texttt{Yejin.Kim@uth.tmc.edu} \\
   \And
   Joydeep Ghosh \\
   Electrical and Computer Engineering\\
   University of Texas at Austin \\
   Austin, TX 78705 \\
   \texttt{jghosh@utexas.edu} \\
}
\begin{document}

\maketitle

\begin{abstract}
\subfile{sections/abstract}

\end{abstract}

\section{Introduction} \label{intro}
\subfile{sections/introduction}

\section{Related Work} \label{lit_review}
\subfile{sections/literature_review}

\section{Background} \label{background}
\subfile{sections/background}

\section{Methodology} \label{methodology}
\subfile{sections/methodology}

\section{Experimental Evaluation} \label{experiments}
\subfile{sections/experiments}

\section{Conclusion} \label{conclusion}
\subfile{sections/conclusion}

\clearpage
\newpage
\small
\bibliography{refs}
\bibliographystyle{plainnat}
\newpage
\appendix
\subfile{sections/appendix}
\clearpage
\newpage
\section*{NeurIPS Paper Checklist}

\begin{enumerate}

\item {\bf Claims}
    \item[] Question: Do the main claims made in the abstract and introduction accurately reflect the paper's contributions and scope?
    \item[] Answer: \answerYes{} 
    \item[] Justification: We support all the claims made in the abstract and introduction by thorough experimental evaluation in the paper. Please refer Section~\ref{experiments} for details.

\item {\bf Limitations}
    \item[] Question: Does the paper discuss the limitations of the work performed by the authors?
    \item[] Answer: \answerYes{} 
    \item[] Justification: Please refer Section~\ref{conclusion}.

\item {\bf Theory Assumptions and Proofs}
    \item[] Question: For each theoretical result, does the paper provide the full set of assumptions and a complete (and correct) proof?
    \item[] Answer: \answerNA{} 
    \item[] Justification: NA.

    \item {\bf Experimental Result Reproducibility}
    \item[] Question: Does the paper fully disclose all the information needed to reproduce the main experimental results of the paper to the extent that it affects the main claims and/or conclusions of the paper (regardless of whether the code and data are provided or not)?
    \item[] Answer: \answerYes{} 
    \item[] Justification: Please refer Section~\ref{experiments} for the experimental details.

\item {\bf Open access to data and code}
    \item[] Question: Does the paper provide open access to the data and code, with sufficient instructions to faithfully reproduce the main experimental results, as described in supplemental material?
    \item[] Answer: \answerNo{} 
    \item[] Justification: The code will be released on acceptance of the paper.

\item {\bf Experimental Setting/Details}
    \item[] Question: Does the paper specify all the training and test details (e.g., data splits, hyperparameters, how they were chosen, type of optimizer, etc.) necessary to understand the results?
    \item[] Answer: \answerYes{} 
    \item[] Justification: Please refer Section~\ref{experiments} for the experimental details.

\item {\bf Experiment Statistical Significance}
    \item[] Question: Does the paper report error bars suitably and correctly defined or other appropriate information about the statistical significance of the experiments?
    \item[] Answer: \answerYes{} 
    \item[] Justification: Please refer Section~\ref{experiments}.

\item {\bf Experiments Compute Resources}
    \item[] Question: For each experiment, does the paper provide sufficient information on the computer resources (type of compute workers, memory, time of execution) needed to reproduce the experiments?
    \item[] Answer: \answerYes{} 
    \item[] Justification: Please refer Section~\ref{experiments} for these details.
    
\item {\bf Code Of Ethics}
    \item[] Question: Does the research conducted in the paper conform, in every respect, with the NeurIPS Code of Ethics \url{https://neurips.cc/public/EthicsGuidelines}?
    \item[] Answer: \answerYes{} 
    \item[] Justification: We understand the NeurIPS Code of Ethics and the work conforms to that.

\item {\bf Broader Impacts}
    \item[] Question: Does the paper discuss both potential positive societal impacts and negative societal impacts of the work performed?
    \item[] Answer: \answerNA{} 
    \item[] Justification: The work in this paper does not have societal impacts.
    
\item {\bf Safeguards}
    \item[] Question: Does the paper describe safeguards that have been put in place for responsible release of data or models that have a high risk for misuse (e.g., pretrained language models, image generators, or scraped datasets)?
    \item[] Answer: \answerNA{} 
    \item[] Justification: NA.

\item {\bf Licenses for existing assets}
    \item[] Question: Are the creators or original owners of assets (e.g., code, data, models), used in the paper, properly credited and are the license and terms of use explicitly mentioned and properly respected?
    \item[] Answer: \answerYes{} 
    \item[] Justification: We have carefully cited the datasets and codes used in the paper.

\item {\bf New Assets}
    \item[] Question: Are new assets introduced in the paper well documented and is the documentation provided alongside the assets?
    \item[] Answer: \answerNA{} 
    \item[] Justification: NA.

\item {\bf Crowdsourcing and Research with Human Subjects}
    \item[] Question: For crowdsourcing experiments and research with human subjects, does the paper include the full text of instructions given to participants and screenshots, if applicable, as well as details about compensation (if any)? 
    \item[] Answer: \answerNA{} 
    \item[] Justification: NA.

\item {\bf Institutional Review Board (IRB) Approvals or Equivalent for Research with Human Subjects}
    \item[] Question: Does the paper describe potential risks incurred by study participants, whether such risks were disclosed to the subjects, and whether Institutional Review Board (IRB) approvals (or an equivalent approval/review based on the requirements of your country or institution) were obtained?
    \item[] Answer: \answerNA{} 
    \item[] Justification: NA.

\end{enumerate}

\end{document}

%% file: sections/abstract.tex
The use of ML systems to estimate individualized treatment effects (ITE) of interventions has facilitated large-scale decision-making in several healthcare applications. However, conventional estimation methods often require substantial amounts of costly  training data for each intervention under consideration. In this work, we present a novel framework, based on causal transformers, for collaborative learning of heterogeneous ITE estimators across distributed institutions (such as hospitals) via Federated Learning. This approach enables training on a large and diverse dataset without the need to share sensitive health data. The proposed method is not only flexible to handle diverse patient populations and non-identical patient measurements (covariates) across different sites, but also allows for the estimation of the treatment effects of disparate treatments being administered across these sites. Moreover, the proposed method can predict the effects of novel and unseen treatments by utilizing the available treatment level information. Thorough experimental evaluation on real-world clinical trial and widely-used research datasets demonstrates that our method surpasses existing baselines. Furthermore, analysis of attention heads learned by our model reveals its proficiency in capturing vital disease and treatment-related information, attesting to the meaningfulness and explainability of the approach.

%% file: sections/introduction.tex
Understanding the impact of an intervention on outcome, also known as the treatment effect, is essential for identifying causation and has broad applications across various disciplines, including healthcare, social sciences, public policy, and more. Since interventions often produce different effects on different individuals, determining the extent to which diverse individuals respond to interventions, known as \emph{Individual Treatment Effects (ITE)} estimation, is an important challenge.

Measuring treatment effects often involves conducting counterfactual analysis, which predicts the outcomes for individuals who have undergone a treatment (factual) under different exposure scenarios (counterfactual) that they haven't actually encountered (and vice versa)~\citep{neyman, rubin}. In recent years, several data-driven machine learning approaches have been extensively utilized for this purpose~\citep{Curth2021NonparametricEO,Curth2021OnIB, insights, HTCE, Assisting_Clinical, MI_bounds, ite_estimation_bounds, bnn}. However, despite these efforts, the application of ITE estimation to randomized controlled trial (RCT) data, often regarded as the gold standard for assessing treatment effects, has been limited. This is primarily due to the constrained availability of RCT data, with much of it being distributed, and data usage agreements that frequently prohibit the export and pooling of patient-level data into a centralized location.

\begin{figure}
\centering
    \centering
    \includegraphics[scale=0.3]{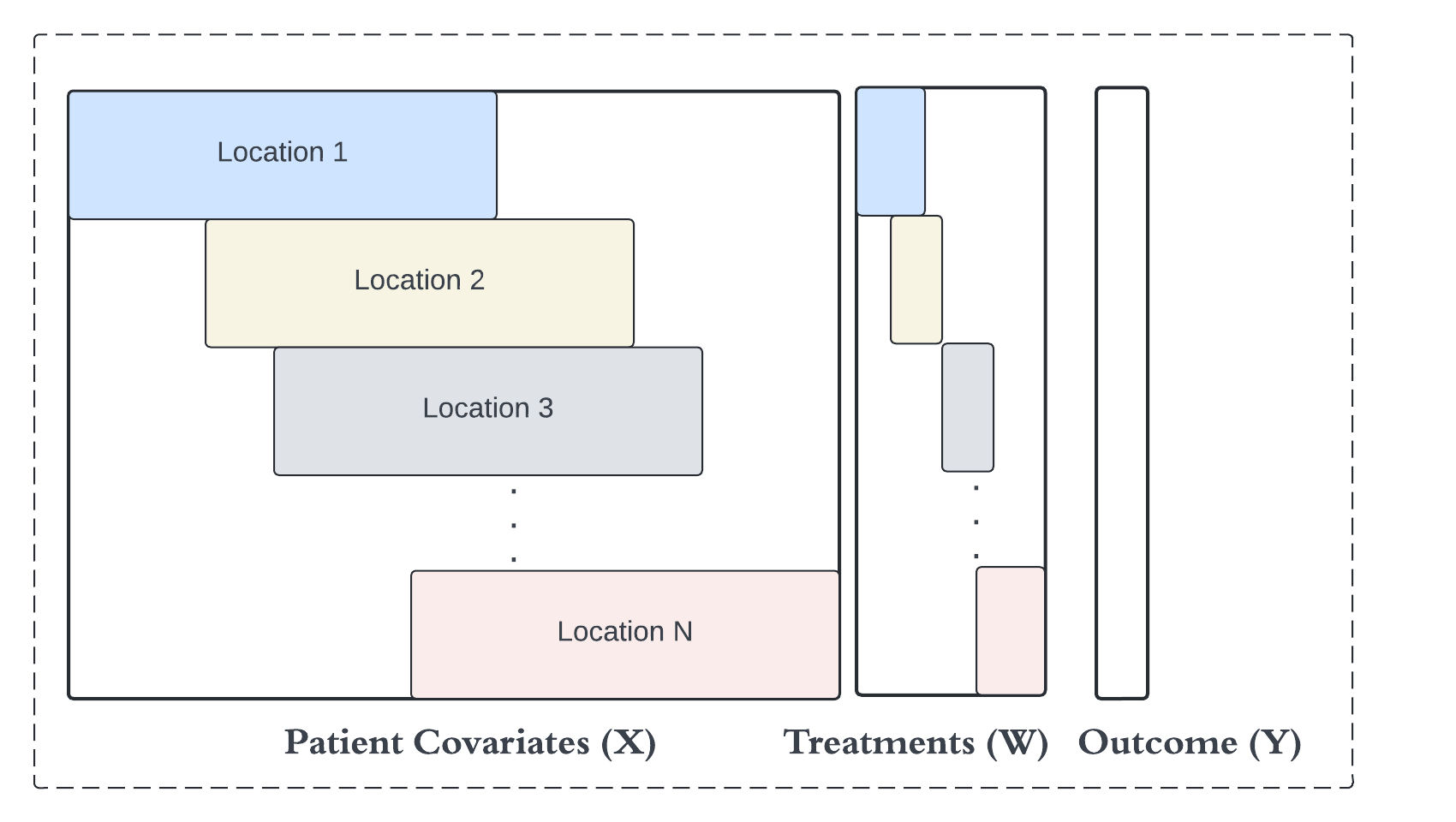}
    \caption{An overview of the setting with N locations, where each location has some overlap with the other locations in their non-identical feature as well as the treatment space.}
    \label{fig:setting}
\end{figure}

This motivates us to investigate Federated Learning (FL)~\citep{fedavg} - a well-known distributed learning paradigm. FL involves learning across numerous organizations or end devices known as clients, each maintaining access to local data while collaborating through a central server. Predictably, FL has been explored for ITE estimation in the past~\cite{federated_CI, bayesian_fl}, particularly focusing on binary treatment scenarios assuming that the data corresponding to the identical treatment (and standard care) is distributed across various sites. But more often than not, the treatments across these sites are disparate with each treatment having distinct characteristics, making them incompatible for aggregated analysis. In addition, some treatments are only observed in specific institutions, rendering the collaborative learning futile for them. This creates a distinct setting depicted in Figure~\ref{fig:setting} where each location corresponding to an institution has non-identical features and treatments. Being able to learn in such a setting would enable the learning method to tap into a wealth of diverse data, including data from previous failed trials, thereby significantly enhancing its learning capacity.

In this work, we propose \emph{Federated Transformers for Treatment Effects Estimation (FedTransTEE)}, an end-to-end framework for ITE estimation across institutions that addresses the challenges of heterogeneous features, treatment, and outcome spaces. Our framework constructs personalized solutions for each institution while also leveraging the data similarities among these institutions to the fullest extent. The proposed framework involves learning a common representation space for the covariates across the participating institutions through a transformer based covariate encoder capable of accommodating non-identical feature sets present across these institutions. The intervention embeddings, that can capture the intervention's features in a representation space, are obtained separately. Both the patient and the intervention embeddings are subsequently fed into another cross-attention transformer, the output of which is utilized by a personalized predictor to estimate the individual treatment effects (ITE) of the unique interventions considered at each site. Furthermore, due to the incorporation of the intervention specific features and/or descriptions, the embeddings of the interventions are generalizable to unseen interventions and can forecast the effects of such newly designed interventions reliably.

To the best of our knowledge, this is the first work that studies the problem of treatment effects estimation across multiple sites under heterogeneous covariate, treatment, and outcome spaces. Specifically, our \textbf{key contributions} can be summarized as :
\begin{enumerate}
    \setlength\itemsep{-0.5em}
    \item We introduce FedTransTEE, a novel causal transformer based model aimed at improving Individual Treatment Effect (ITE) estimation. FedTransTEE is designed to accommodate heterogeneous covariates, treatments, and outcomes spaces.
    \item We propose a federated learning framework based on FedTransTEE, enabling it to leverage distributed and disparate datasets across multiple sites.
    \item Extending the framework to the zero-shot ITE estimation setting, we demonstrate its effectiveness in estimating the effects of novel and unseen treatments reliably.
    \item Since the proposed novel framework is capable of operating in heterogeneous feature spaces, it also advances the field of Federated Learning (FL) methods, enhancing their utility in diverse settings.
    \item A thorough evaluation of the proposed method is performed on real-world clinical-trial and research datasets highlighting its efficacy in ITE estimation as well as collaborative learning from multiple sites. Additionally, a model interpretability approach based on examining the attention heads learned by the proposed method demonstrates the explainability of the learned latent representations, making it valuable for diagnostic applications. 
\end{enumerate}

%% file: sections/literature_review.tex
This section provides a brief overview of the most relevant prior work in the fields of ITE estimation, federated learning and federated learning for healthcare and ITE estimation.

\paragraph{ITE Estimation} Due to their ability to handle high-dimensional data and intricate feature interactions, machine learning methods have been increasingly used for ITE estimation. However, in the majority of the cases, the estimation of effects is typically conducted directly by using locally accessible data sources. The ITE estimation methods can broadly be divided into two categories - ones that estimate the potential outcomes of each intervention and use the difference between the estimated outcome with the intervention and the estimated outcome without the intervention to obtain the ITE called the indirect learners, and the other methods that try to directly model the ITE called the direct learners. Amongst the methods that predict the outcomes for each intervention separately, some methods augment the covariates by adding the intervention information to create a set of independent variables $(\Xx, \T)$ that are fed to the model to predict the $\mu_1(x)$ and $\mu_0(x)$ like the methods using regression trees~\citep{regression_trees}, random forests~\citep{Knzel2017MetalearnersFE}, non-parametric methods~\citep{Curth2021NonparametricEO}, and Bayesian methods~\citep{bayesian_hill}. But due to the selection bias associated with intervention assignment to individuals, conventional supervised learning methods that use intervention as a feature might not be directly applicable to the task of estimation. Therefore, other set of works focused on building separate models, one for each potential outcome, under the Neyman-Rubin framework. The earlier works in this area used techniques like regression based modeling~\citep{Cai2011AnalysisOR}, trees, random forests~\citep{Athey2015MachineLM, Foster2011SubgroupIF}, and others~\citep{Nie2017QuasioracleEO, doubly_robust}. Then recently, the focus has shifted on using neural networks where the most common approach is to segregate the estimation into learning a representation from the input and then building intervention specific predictive models on top of the representation~\citep{Curth2021OnIB, deep_treat, bnn, pmlr-v129-qidong20a, ite_estimation_bounds, bayesian_ite_gp, Zhang2020LearningOR, bica2022transfer, Hassanpour2020Learning, NEURIPS2018_a50abba8, Hassanpour2020Learning, curth2021really, hyper-ite}. On the other hand, the direct learners often estimate a set of parameters called the nuisance parameters first and then use these to directly estimate the treatment effects, different direct learners having different nuisance parameters and unique methods for learning these nuisance parameters~\citep{Wager2015EstimationAI, Powers2017SomeMF, ganite, Kristiadi2019UncertaintyQW}.

\paragraph{Federated Learning} Initially presented as the FedAvg algorithm in the pioneering study by~\citep{fedavg}, FL has since undergone numerous modifications tailored to address distinct challenges. These adaptations include both global solutions as well as personalized solutions that cater to the non-IID data across clients in a better way. Some key works that obtain global solutions for FL include~\citep{fedrep, scaffold, fedpd, feddyn, fedbe} and that for personalised FL include~\citep{fedprox, pfedme, Makhija2022ArchitectureAF, personalised_meta_learning}. However, it is only very recently that the problem of heterogeneous feature spaces has been explored in the FL setting.~\citet{suzuki2023clustered} use clustering to figure out similar clients and exchange knowledge within the cluster whereas~\citet{rakotomamonjy2023personalised} use learnt prototypes to align the feature spaces across the clients. 

\paragraph{Federated Learning for Healthcare and ITE Estimation} Given FL's recognized utility for  healthcare~\citep{Rieke2020TheFO, Sheller2020FederatedLI, Prayitno2021ASR}, numerous applications have leveraged FL for various tasks in the healthcare domain like predicting adverse drug reactions~\citep{fl_adverse_drug_reactions}, stroke prevention~\citep{Ju2020PrivacyPreservingTT}, mortality prediction~\citep{mortality_prediction}, etc. But limited attention has been devoted to the federated estimation of causal effects.~\citet{federated_CI} proposed a method to predict the average treatment effects by first locally computing summary statistics and then appropriately aggregating these statistics across sites.~\citet{bayesian_fl} use a Bayesian mechanism that integrates local causal effects from different sites to estimate posterior distributions of the effects but does not allow for non-identical data distributions across sites, and~\citet{vo2} then extended it to dissimilar data distributions across sites.~\citet{face, han2023privacypreserving} proposed privacy preserving learning across sites but assumes presence of identical covariates which is often not the case in practice.~\citep{, Yang2018CombiningMO, Zeng2023EfficientGA, han2023multiply} consider non-identical covariates across sites but solve the problem using transportability by trying to transport the causal effects from a source population distributed across multiple sites to a target population. While some of these methods consider heterogeneous covariates across sites, they do not consider disparate interventions or treatments being administered across sites.

%% file: sections/background.tex
The ITE estimation problem refers to prediction of the effects of different interventions on individual subjects. A specific intervention data is denoted as $\D = (\Xx, \Y, \T)$, where $\Xx$ encodes the pre-intervention covariate vectors of the subjects, $\T$ encodes the intervention (or treatmment), and $\Y$ denotes the outcome corresponding to the intervention. Without loss of generality, we consider a $K$ intervention setting, wherein $\T \in \{0, 1, \dots K\}$ with $\T = 0$ denoting no intervention and $\T = j$ denoting use of the $j^{th}$ intervention. We use notation $\Y(0)$ to record the outcome under placebo or standard-of-care intervention and $\Y(j)$ to record the outcome under the $j^{th}$ intervention. For each subject in the population, we only see one of the outcomes associated with the intervention that was used on the subject. The probability of the intervention assignment or the propensity score is denoted by $\pi(x,j) = \mathbb{P}(\T = j | \Xx = x)$. The Neyman-Rubin~\cite{neyman, rubin} framework for causal inference suggests designing a separate potential outcome function for each intervention that can be applied on individual subjects, $\mu_j(x) = \mathbb{E}(\Y | \Xx = x, \T = j )$. These functions are then used to predict the effect of the treatment $j$. 

The framework operates under three key \textit{assumptions} (which our method and the baselines also rely on). The first of these assumptions \textit{Stable Unit Treatment Value Assumption (SUTVA)} requires that each individual's potential outcomes are independent of the potential outcomes of other individuals and that there is no interference across individuals. The second assumption is called \textit{unconfoundedness} and suggests that there are no hidden confounders affecting both the treatment assignment and the outcome, i.e., the potential outcomes, $\mu_j(x)$, are independent of the treatment variable, $\T$, given the observed variables $\Xx$. And the last assumption requires the treatment assignment policy to be stochastic such that each individual has a non-zero probability of assignment for each treatment. Under these assumptions, the ITE of an intervention $j$ is approximated by the Conditional Average Treatment Effect (CATE), $\tau_j(x) = \mu_j(x) - \mu_0(x)$. A learning model tries to estimate the CATE by predicting $\hat{\mu}_j(x)$ and $\hat{\mu}_0(x)$, and its estimates are denoted by $\hat{\tau}_j(x) = \hat{\mu}_j(x) - \hat{\mu}_0(x)$.  

Now, consider an FL setting with $N$ clients (institutions or hospital locations in this case). Each client $m$ collects the local dataset $\D_m = (\Xx^m, \Y^m, \T^m)$ corresponding to its local population, intervention and associated outcomes. We denote the number of subjects on the client $m$ by $n_m$ and the corresponding covariate vectors to be encoded in $d_m$ dimensional-space, i.e., $\Xx^m \in \mathbb{R}^{n_m \times d_m}$. Further, we consider multiple interventions per client, and without loss of generality assume client $m$ to be delivering interventions denoted by $\T^m$, and recording the factual outcomes $\Y^m$. Also note that since the population distribution across the clients is generally non-IID, $\mathbb{P}(\Xx^m) \neq \mathbb{P}(\Xx^l)$ for any two locations $m \neq l$, this is the innate heterogeneity in the data across clients which is further exacerbated by the disparate interventions $\T^m \neq \T^l$, and different subsets of variables accessible on each location $d^m \neq d^l$. The outcomes ($\Y^m$) across different locations can be identical or different.

%% file: sections/methodology.tex
\begin{figure}
\centering
    \centering
    \includegraphics[scale=0.4]{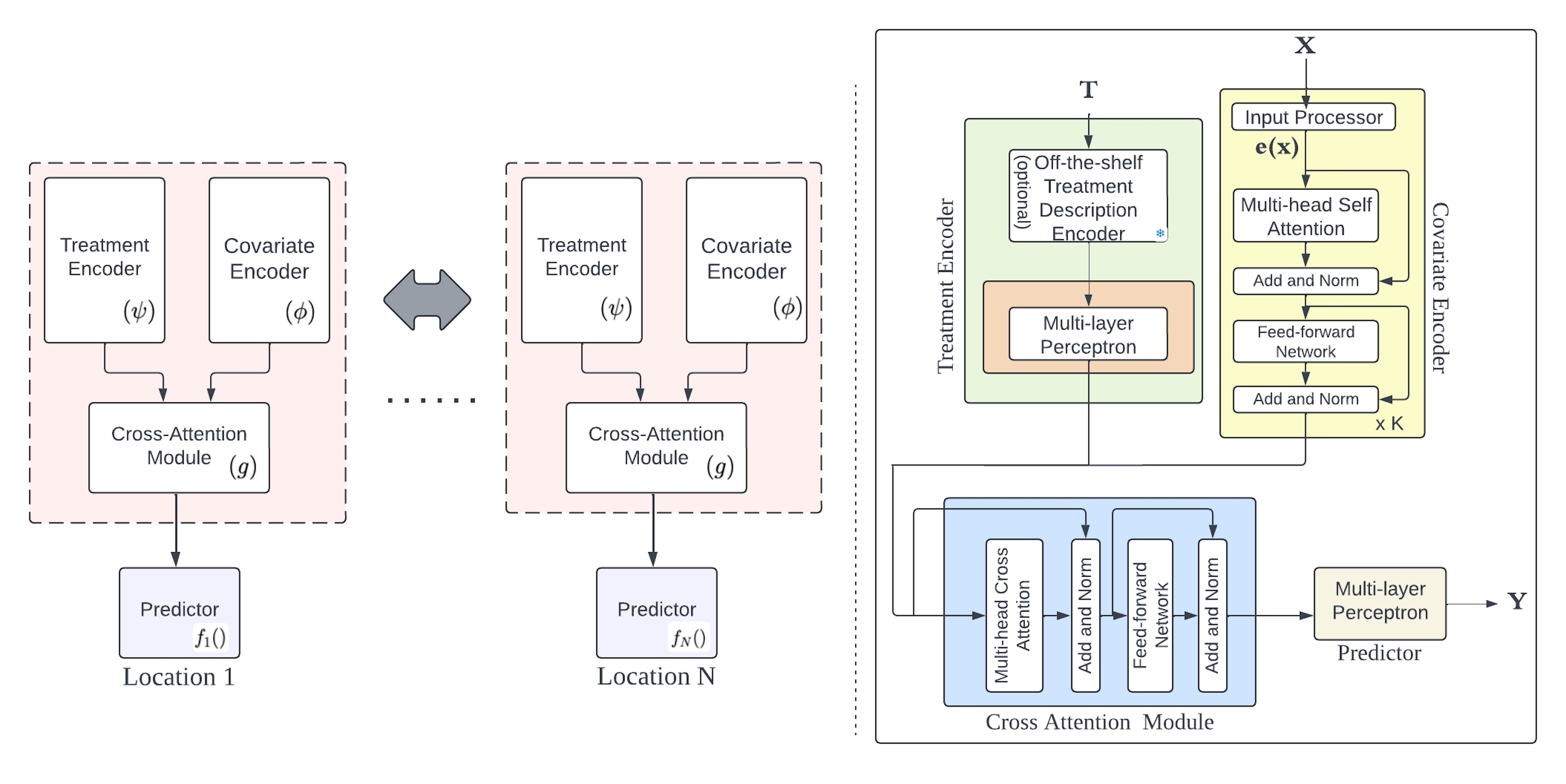}
    \caption{The above figure illustrates the FedTransTEE framework, with $N$ site locations on the left; as depicted, the covariate encoder, the treatment encoder and the cross-attention module are shared across locations but the predictor is personalised for each location. A detailed view of the model architecture with specifics of each component is shown on the right side. 
    }
    \label{fig:method}
\end{figure}

In this section we present the details of the proposed framework - FedTransTEE. The right side of Figure~\ref{fig:method} shows the workflow of our method used at each site, which includes the following components - i) the covariate encoder that takes the individual subject (patient) features to create a subject-level latent representation (embedding), ii) the treatment encoder that encodes the specific intervention into another latent representation, iii) the cross attention module that models the interaction between the covariate embedding and the treatment embedding, and iv) a predictor that is used to predict the outcome of the treatment. Our framework is motivated by randomized clinical trials, yet it is designed to operate within the context of observational data, where treatment assignment is non-randomized. In this paper, we explicitly demonstrate the framework for the problem of federated ITE estimation in healthcare settings, and interchangeably use the terms clients and sites, features and covariates, and patients and subjects. Next we will present the details of the each module.

\subsection{Covariate encoder} At different healthcare sites, patient data is recorded using varied protocols, leading to differences in the types of features collected and how these features are represented. To address this variability, we've designed a covariate encoder that can work with these diverse feature sets for collaboratively learning across sites. Transformers have exhibited remarkable proficiency in learning representations and processing inputs of varying lengths, making them well-suited as the core module for representation learning in the covariate encoder. Each patient covariate vector undergoes processing (discussed in next section) to generate a sequence of tokens which is fed through a learnable embedder to convert it into a sequence of 256-dimensional embeddings. This sequence of embeddings then proceed through a transformer containing $k$ layers of multi-headed self-attention modules. The entire transformer module enables the acquisition of an effective patient representation, which is then outputted by the covariate encoder. The input-processing unit and the transformer module are collectively known as the covariate encoder and are denoted by $\phi$. Below, we specify the particulars of the input processing and the transformer based representation learning.

\textbf{Input Processing} The input processing module takes in the names and values of each covariate for every patient along with the specifications indicating whether the information in each covariate is numerical or categorical and outputs a joint embedded vector (containing the sequence of embeddings) that includes all the patient level features. If $\vecX = [x_1, x_2, \dots x_{d_m}]$ is a covariate vector at the $m^{th}$ site, the processed vector, $\vecX' = f(\vecX) = [h_1(x_1), h_2(x_2), \dots h_{d_m}(x_{d_m})]$, is obtained by applying function $h_i : x_i \rightarrow \mathbb{R}^d$ on each of the $x_i$'s. If $x_i$ is a categorical feature, the name of the feature is concatenated with the value of the feature to create a sequence of tokens, which is then tokenized and passed through an embedder to get embeddings. If $x_i$ is a numerical feature, the embedding is obtained for the name of the feature and the embedding is multiplied by the value of $x_i$ to generate the final embedding. The embedder employs a learnable 256-dimensional embedding for each token in the vocabulary and is accessed via a lookup table. Each feature's embedding is then concatenated to obtain $\vecX'$ and a special learnable [CLS] token is prepended to $\vecX'$ to obtain the final embedding of $\vecX$, \textbf{e(x)}  = [[CLS] $, h_1(x_1), h_2(x_2), \dots h_{d_m}(x_{d_m})]$. This method of processing the input helps in capturing the semantic meaning of the covariates and is also seen in other works that use transformers on tabular data like in~\citep{wang2022transtab, xtab}.

\textbf{Representation Learning} The initial sequence of embeddings, \textbf{e(x)}, of the covariates are then passed through a transformer network consisting of $k$-layers where each layer is composed of a multi-head self attention module followed by a multi-layer perceptron with residual connection in between and layer normalization at the end. The attention mechanisms encoded in the attention matrix allows the model to focus on different parts of the inputs for learning a better representation and multiple attention heads are used to allow it to attend to different parts of the input sequence simultaneously. We use a 8-head self-attention network in our architecture.

\subsection{Treatment Encoder} 
The treatment encoder ($\psi$) is used to convert specific treatments into 256-dimensional embeddings, denoted by $\psi(j)$ for treatment $j$. This component includes an optional treatment information encoder, which processes information like textual descriptions or drug compositions about the treatments to create a representation of the treatment in a high-dimensional space. Following this, there's a learnable two-layer MLP (multi-layer perceptron) with a ReLU activation function in between. If additional details about a treatment are provided, the details are fed into the information encoder and its output is given to the MLP to generate the final 256-dimensional treatment embedding. Alternatively, if no additional information is available, the MLP takes a one-hot representation of the 1-of-$K$ treatments used on specific subjects as input to generate the final embedding. In our experiments (wherever mentioned), we consider the textual descriptions of the treatments from clinicaltrials.gov website as the supplementary information, and utilize pre-trained language models like GPT, BERT, etc. for encoding these descriptions. Other specialized off-the-shelf or pre-trained encoders capable of interpreting specific treatment level information can be plugged in for use in our framework. This treatment information encoder proves particularly valuable in zero-shot scenarios, as it predicts the outcomes of newly introduced and distinct treatments by leveraging existing similarities with treatments already seen by the model, as demonstrated in our experiments later on.

\subsection{Cross Attention Module and Predictor} 
The cross-attention module, $g()$, takes in the treatment and the patient embeddings and learns the interaction between them through a cross-attention transformer block. As opposed to self-attention, the cross attention uses the treatment embedding as the query and the sequence of feature embeddings in the patient embedding as the keys and values. Multiple cross-attentions are employed to facilitate information exchange between parallel treatment and patient representations. We use a transformer block of single layer with 8 attention heads for the cross-attention encoder. Finally, a predictor constructed from a two-layer MLP with ReLU activation non-linearity is used to predict the observed (factual) outcome for the given treatment and the covariate vector, $\hat{y} = f_m (g(\psi(j),\phi(x)))$ for the $m^{th}$ site. Employing a separate predictor allows for personalized modeling within the FL framework. This approach enables collaborative learning of joint treatment and patient embeddings across all sites, while keeping the prediction head independent to handle diverse outcomes measured across these sites.

\subsection{Local Optimization and Collaboration} 
The overall training process consists of $E$ communication rounds between the clients and $e$ local training epochs on individual clients. In each communication round, the globally aggregated covariate encoder ($\bar{\phi}$), treatment encoder ($\bar{\psi}$), and the cross-attention module ($\bar{g}$) are broadcasted to all the clients. The clients initialize their corresponding local modules with the obtained global modules and train the entire local model for $e$ epochs, at the end of which the locally trained models are sent to the server where they are again aggregated.

The local optimization at each client involves an alternate minimization procedure where the predictor and the representation learning modules are optimized alternatively on the mean squared error loss between the prediction and the true outcome of the treatment on the given patient. Specifically, at each client $m$, the optimization procedure first updates the predictor $f_m$ by
$$
f_m^* = \argmin_{f} \E_{(x,t,y) \sim \D_m} \bigg[  \bigg( f (\bar{g}(\bar{\psi}(t), \bar{\phi}(x))) - y \bigg)^2  \bigg],
$$
and then uses the optimized predictor to learn the other components
$$
\{\psi_m^*, \phi_m^*, g_m^*\} = \argmin_{\{\psi_m, \phi_m, g_m\}} \E_{(x,t,y) \sim \D_m} \bigg[ \bigg( f_m^* (g_m(\psi_m(t),\phi_m(x))) - y \bigg)^2  \bigg].
$$
The optimized $\{\psi_m^*, \phi_m^*, g_m^*\} $ are uploaded on the server and aggregated parameter-wise where the parameters of the clients are weighted according to the number of data points present on each client. The global aggregated versions of the models are obtained in the following way -
$$
\bar{\psi} = \sum_{i=1}^{N} \dfrac{n_i}{\sum_{i'=1}^{N} n_{i'}} \psi_i ; \quad \bar{\phi} = \sum_{i=1}^{N} \dfrac{n_i}{\sum_{i'=1}^{N} n_{i'}} \phi_i ; \quad \bar{g} = \sum_{i=1}^{N} \dfrac{n_i}{\sum_{i'=1}^{N} n_{i'}} g_i.
$$
The aggregated parameters are sent to the clients again for the next round of training, and the entire procedure continues for a total $E$ number of communication rounds.

%% file: sections/experiments.tex
In this section, we demonstrate the experimental evaluation of our method and compare it with various baseline approaches. Additionally, we also aim to explore additional questions related to the zero-shot learning capability of the framework, and the interpretability of the approach.
\subsection{Experimental Setting}
\textbf{Datasets} For the first set of experiments, we consider three prevalent semi-synthetic datasets in the ITE research, corresponding to the Infant Health and Development Program (IHDP) dataset~\cite{ite_estimation_bounds}, 2016 Atlantic Causal Inference Conference (ACIC-2016) Competition dataset~\cite{acic16} and Twins~\cite{twins} dataset. All the three datasets have binary treatments. The experimentation protocol based on semi-synthetic datasets is used for two reasons -  i) the presence of ground-truth data for factual and counterfactual outcomes enables calculation of treatment effects and evaluation across various metrics such as PEHE, ATE, etc., and ii) their binary treatment setting renders them suitable for baseline comparisons, particularly for federated learning methods that exclusively deal with binary treatments. We also obtain two real-world datasets - i) data collected from three randomized clinical trials for intracerebral hemorrhage (ICH) therapy development: ATACHII (NCT01176565), MISTIEIII (NCT01827046) and ERICH (NCT01202864)~\cite{ich_dataset}, and ii) the CPAD dataset~\cite{cpad_dataset} containing Alzheimer's disease or mild cognitive impairment from 38 Phase II/III randomized clinical trial data. More details and the statistics of datasets are included in Appendix~\ref{app:datasets} and Table~\ref{tab:dataset_desc} due to space constraints.

\textbf{Baselines} For the binary treatment datasets obtained from the open-source, we compare our method against three different types of baselines - the methods that can work in FL setting by estimating ITE from multiple sources of data, namely iFedTree~\cite{ifedtree} and FedCI~\cite{federated_CI}, note that iFedTree only predicts the factual outcomes. Then we compare our method against the popular and latest ITE estimation methods that do not work with multiple sources using CATENets package~\cite{catenets}. We train them in the following ways - i) local training by training individual models locally on each site without any collaboration (because there is no way of collaborating across sites in these methods), ii) centralized training by collecting all the data from different sources on a single site and training on the entire dataset. Note that the centralized training  is only for a hypothetical comparison as it violates the individual patient data privacy.

\textbf{Multi-site Simulation} In order to simulate the decentralized federated learning setting for testing our method and the baselines, we randomly partition the semi-synthetic datasets into $N$ clients. The IHDP dataset is partitioned into 3 sites, the ACIC-16 into 5 sites and the Twins dataset into 10 sites, with each site containing 240, 960 and 1140 datapoints respectively. The partitions are made heterogeneous by varying the ratios of the patients with treatment and without treatment across sites. Among the real-world datasets, the ICH dataset is divided into three parts, each linked to a specific hospital location where a unique treatment was implemented. Similarly, the CPAD dataset encompasses 19 sites, reflecting data gathered from 19 distinct study locations. The local data partition for each client is segmented into training, validation, and test sets through a 70:15:15 split. To assess performance, the test dataset from the specific site is utilized.

\textbf{Metrics} Given the unavailability of ground-truth treatment effects in real-world data, directly evaluating ITE estimation methods is challenging. Consequently, we employ two distinct sets of metrics for experiments conducted on semi-synthetic datasets and real-world datasets. With the semi-synthetic datasets where both the factual and the counterfactual outcomes are available, we measure and report the Root Mean Squared Error on the factual outcome (RMSE-F), the Error in the Average Treatment Effect (ATE$_\epsilon$), and the Precision in the Estimation of Heterogeneous Effects (PEHE). And for the real-world datasets when only the factual outcome is available we report the RMSE-F, and the difference in Average Treatment on Treated (ATT$_{\epsilon}$). Due to the space limitation, the definitions of the metrics are included in Appendix~\ref{app:metrics}. 

\textbf{Training parameters} The experiments are conducted over 5 rounds, with performance metrics evaluated on a held-out test dataset. For centralized experiment settings, the entire dataset is partitioned into a train, validation and test split with corresponding ratios 70:15:15. As mentioned, our covariate encoder comprises 2 transformer layers, each consisting of 8 attention heads, while the cross-attention module includes one transformer layer with 8 attention heads, all operating within a hidden dimensionality of size 256. For the off-the-shelf treatment description enocder module in the treatment encoder, we employ the BERT model~\cite{devlin2019bert} to generate a 786-dimensional embedding of the textual descriptions of the treatments. In absence of the treatment descriptions, we use a one-hot encoding to initially embed the treatments. The entire procedure is configured to run for 200 communication rounds, with each site undergoing 5 local training epochs and an early stopping patience of 20 epochs. Adam optimizer is utilized for training the model, with a learning rate set to $5e-3$ and a batch size of 128. All the hyper-parameters of the training procedure for our method and the baselines are tuned on a set-aside validation set, and all the models are trained on a 4 GPU machine with GeForce RTX 3090 GPUs and 24GB per GPU memory.

\subsection{Results}
The performance of our method and the baselines on semi-synthetic and real-world datasets is reported in Table~\ref{table:performance_syn} and Table~\ref{table:performance_real} respectively. Firstly, we observe in Table~\ref{table:performance_syn} that the proposed method, FedTransTEE in FL setting, demonstrates superior performance compared to federated baselines. The better performance of our approach can be attributed to the innovative and highly effective collaboration it facilitates. Secondly, if the data were centralized (although typically not possible due to privacy concerns), the proposed method outperforms all the other methods suggesting its efficiency in predicting ITE on the collective data source. Subsequently, to demonstrate the real-world setting for the real-world datasets (distributed data sources and centralization impossible), we show results in Table~\ref{table:performance_real}. In this scenario, the single-source baselines have to operate in the local learning setting and other FL baselines are inapplicable due to their inability to accommodate multiple treatments across sites. Notably, our method not only exhibits superior average performance but also demonstrates lower variance. This is due to the fact that when there are multiple sites with diverse data and different amounts of data sources, the collaboration can help in reducing the discrepancy between the performance across clients, providing robust ITE estimation across all the sites. Additional results corresponding to the local training of the models are included in the Appendix~\ref{app:experiments} due to the space limitation.

\begin{table*}[!ht]
\centering
\caption{The table shows performance comparison between our method and the related ITE prediction methods on the held-out test dataset for the semi-synthetic datasets averaged over 5 runs.}
\label{table:performance_syn}
\setlength\tabcolsep{3pt}
\begin{center}
\begin{small}
\scalebox{0.8}{
\hspace*{-1.5cm}
\begin{tabular}{@{}lccccccccc@{}}
\toprule
 \multirow{2}{*}[-0.5ex]{Method} & \multicolumn{3}{c}{IHDP} & \multicolumn{3}{c}{ACIC-16} & 
 \multicolumn{3}{c}{Twins} \\
   \cmidrule(lr){2-4} \cmidrule(lr){5-7} \cmidrule(lr){8-10}
        & (PEHE) & (ATE$_\epsilon$) & (RMSE-F) & (PEHE) & (ATE$_\epsilon$) & (RMSE-F) & (PEHE) & (ATE$_\epsilon$) & (RMSE-F) \\[0.5ex]
 \midrule
 FedCI & 1.33 $\pm$ 0.20 & 0.65 $\pm$ 0.14 & 2.59 $\pm$ 0.09 & 2.3$\pm$0.04  &  1.14$\pm$0.19  & 1.64$\pm$0.02  & 0.34 $\pm$ 0.01 & 0.052 $\pm$ 0.01 &  0.56 $\pm$ 0.1 \\ [0.3ex]
 iFedTree &  -  & - &  2.0$\pm$0.15 &   - &   - &  2.53$\pm$0.1  &  -  &  -  &  0.09$\pm$0.001 \\ [0.3ex]
 FedTransTEE (Ours) & \textbf{1.02 $\pm$ 0.01} & \textbf{0.26 $\pm$ 0.06} & \textbf{1.77 $\pm$ 0.5} &  \textbf{0.78 $\pm$ 0.1} &  \textbf{0.326 $\pm$ 0.02} & \textbf{0.728 $\pm$ 0.01} & \textbf{0.32 $\pm$ 0.01} & \textbf{0.01 $\pm$ 0.002} &  \textbf{0.09$\pm$0.01} \\ [0.3ex]
 \midrule
 (Central) & & & & & & & & & \\ [0.3ex]
 S-Learner$_{c}$ & 0.93 $\pm$ 0.005 & 3.7 $\pm$ 0.001 & 1.22 $\pm$ 0.02 &  2.6 $\pm$ 0.009 &  3.4 $\pm$ 0.002 & 0.61 $\pm$ 0.003 & 0.32 $\pm$ 0.01  & 0.01 $\pm$ 0.001  & 0.08 $\pm$ 0.02 \\ [0.3ex]
 T-Learner$_{c}$ & 1.22 $\pm$ 0.02  & 3.6 $\pm$ 0.01 & 1.23 $\pm$ 0.04 & 3.8 $\pm$ 0.01 &  3.46 $\pm$ 0.001 & 0.56 $\pm$ 0.005 & 0.33 $\pm$ 0.01 & 0.02 $\pm$ 0.001  &   0.09 $\pm$ 0.003 \\ [0.3ex]
 TARNet$_{c}$ & 1.19 $\pm$ 0.002 & 3.9 $\pm$ 0.01 &  1.26 $\pm$ 0.02 &  2.94 $\pm$ 0.01 &  3.54 $\pm$ 0.001 &  0.41 $\pm$ 0.001 & 0.32 $\pm$ 0.001 & 0.02 $\pm$ 0.001  &   0.09 $\pm$ 0.005 \\ [0.3ex]
 FlexTENet$_{c}$ & 1.19 $\pm$ 0.005 & 3.9 $\pm$ 0.05 & 1.2 $\pm$ 0.01 &  2.83 $\pm$ 0.01 &  3.5 $\pm$ 0.001 & \textbf{0.34 $\pm$ 0.001} & 0.32 $\pm$ 0.001 & 0.015 $\pm$ 0.001 &   0.085 $\pm$ 0.001 \\ [0.3ex]
 HyperITE$_{c}$ (S-Learner) & 0.91 $\pm$ 0.001 & 3.9 $\pm$ 0.01 & 1.19 $\pm$ 0.02 &  2.6 $\pm$ 0.01  &  3.5 $\pm$ 0.01 & 0.44 $\pm$ 0.01  &  0.32 $\pm$ 0.002  &  0.02 $\pm$ 0.001  &  0.085 $\pm$ 0.01 \\ [0.3ex]
 HyperITE$_{c}$ (TARNet) & 1.08 $\pm$ 0.04 & 3.8 $\pm$ 0.003 & 1.26 $\pm$ 0.009 &  2.9 $\pm$ 0.02  &  3.5 $\pm$ 0.02  &  0.6 $\pm$ 0.01  &  0.32 $\pm$ 0.01 & 0.02 $\pm$ 0.001 & 0.089 $\pm$ 0.02 \\ [0.3ex]
 FedTransTEE$_{c}$ (Ours) & \textbf{0.90 $\pm$ 0.01} & \textbf{0.14 $\pm$ 0.1} & \textbf{0.97 $\pm$ 0.01} &  \textbf{0.78 $\pm$ 0.01} & \textbf{0.37 $\pm$ 0.10}  &  0.45 $\pm$ 0.04  &  \textbf{0.30 $\pm$ 0.01 } &  \textbf{0.01 $\pm$ 0.004} &  \textbf{0.08 $\pm$ 0.002} \\[0.3ex]
\bottomrule
\end{tabular}
}
\end{small}
\end{center}    
\end{table*}

\begin{table*}[!ht]
\centering
\caption{The table presents a performance comparison between our method and related ITE prediction methods for real-world datasets. The reported metrics in these results are averaged across all clients for each run.}
\label{table:performance_real}
\setlength\tabcolsep{3pt}
\begin{center}
\begin{small}
\scalebox{0.8}{
\hspace*{-1.5cm}
\begin{tabular}{@{}lcccc@{}}
\toprule
\multirow{2}{*}[-0.5ex]{Method} & \multicolumn{2}{c}{ICH} & \multicolumn{2}{c}{CPAD} \\
   \cmidrule(lr){2-3} \cmidrule(lr){4-5} 
        & (RMSE-F) & (ATT$_\epsilon$) & (RMSE-F) & (ATT$_\epsilon$)  \\[0.5ex] 
    \midrule
 S-Learner$_{l}$ & 1.32 $\pm$ 0.35  & 0.22 $\pm$ 0.01 &  7.4 $\pm$ 3.8  &  10.1 $\pm$ 0.01  \\ [0.3ex]
 T-Learner$_{l}$ &  1.34$\pm$0.41  &  0.24 $\pm$ 0.02 &  6.9 $\pm$ 3.2  &  9.8 $\pm$ 0.02 \\ [0.3ex]
 TARNet$_{l}$ &  1.33 $\pm$ 0.44  &  0.23 $\pm$ 0.01 &  7.1 $\pm$ 3.04  &  9.5 $\pm$ 0.01 \\ [0.3ex]
 FlexTENet$_{l}$ &  1.33 $\pm$ 0.38  &  0.23 $\pm$ 0.03 &  5.6 $\pm$ 0.45  &  \textbf{8.1 $\pm$ 0.89} \\ [0.3ex]
 HyperITE$_{l}$ (S-Learner) &  1.30 $\pm$ 0.32  &  0.22 $\pm$ 0.01 &  7.45 $\pm$ 3.58  &  13.4 $\pm$ 1.6 \\ [0.3ex]
 HyperITE$_{l}$ (TARNet) &  1.32 $\pm$ 0.28  &  0.22 $\pm$ 0.01 &  7.2$ \pm$ 2.7  &  11.2 $\pm$ 2.1 \\ [0.3ex]
 FedTransTEE (Ours) & \textbf{1.19 $\pm$ 0.09}  &  \textbf{0.10 $\pm$ 0.02}  &  \textbf{4.6 $\pm$ 0.06}  &  10.2 $\pm$ 0.02 \\ [0.3ex]
\bottomrule
\end{tabular}
}
\end{small}
\end{center}    
\end{table*}

\subsubsection{Zero-shot ITE Estimation}\label{subsec:zero_shot}
We further evaluate the performance of the proposed method in zero-shot testing scenarios. This entails situations where there is no historical data available for a newly designed treatment, or when a new site that seeks to estimate the effects of its treatments is introduced. In such cases, we explore how the proposed framework can be utilized for estimating the effects of the unseen treatments. To enable zero-shot learning, we obtain additional information like a description (or a set of features) for the unseen treatment, and use this information to generate an embedding of the treatment, and pass it to our proposed framework (which was trained on the data and descriptions of the other treatments). To assess this capability, we set up an experiment where we exclude data related to a specific treatment (ATACH-II or MISTIE-III) from the ICH training dataset. We then train the model solely on data from the other two treatments, using treatment descriptions obtained from clinicaltrials.gov (excerpts of which are shown in Table~\ref{tab:treatment_desc}). The model is subsequently tested on the data corresponding to the excluded treatment (ATACH-II or MISTIE-III). The results of this zero-shot performance evaluation are presented in Table~\ref{table:zero_shot}. For comparison, we also include metrics for the same test dataset when it was part of the training procedure, shown under the supervised performance column in Table~\ref{table:zero_shot}.

\begin{table*}[!ht]
\centering
\caption{Test RMSE-F on ICH dataset under the \textbf{zero-shot} training protocol.}
\label{table:zero_shot}
\setlength\tabcolsep{3pt}
\begin{center}
\begin{small}
\scalebox{0.8}{
\hspace*{-1.5cm}
\begin{tabular}{@{}lccc@{}}
\toprule
 Treatment Name & Supervised Performance & Zero-shot Performance & $\Delta$ \\ [0.3ex]
 \midrule
 ATACH-II & 1.21$\pm$0.01 & 1.30$\pm$0.06 & $\sim$0.1 \\ [0.3ex]
 MISTIE-III & 0.82$\pm$0.05 & 1.02$\pm$0.04 & $\sim$0.2 \\ [0.3ex]
\bottomrule
\end{tabular}
}
\end{small}
\end{center}    
\end{table*}

\subsubsection{Interpretability Analysis}\label{subsec:interpret}
\textbf{Important covariates for outcome prediction.} We investigate the self-attention mechanism of the covariate encoder to identify interpretable patterns in the relationships among covariates for predicting outcomes in ICH therapy trials. By examining the attention weights, we can discern which features the model deems important for its predictions. The attention weights are obtained by averaging the attention activations over all patients. In ICH data, four self-attention heads had distinguishable attention weight patterns.
Attention Head 1 revealed that the final embedding ([CLS] token) heavily attends to a patient's prior history, such as racial group (being White or not) and type 2 diabetes. This finding aligns with established knowledge that race and diabetes comorbidity significantly impact clinical outcomes, such as functional independence measured by mRS score after ICH onset~\cite{10.1001/jamanetworkopen.2022.1103,ZHENG2018e756}.
Attention Head 2 showed a strong connection between prior vascular conditions and hematoma pathology. Key variables, including the final embedding, intraventricular hemorrhage volume, anticoagulant use, Asian race, and prior history of heart failure, were linked to hematoma volume. The final embedding also attends to the location of the hematoma. This indicates that Attention Head 2 captures the pathology of hematoma development and progression, a direct therapy target for ICH.
Attention Head 3 demonstrated a clear connection between the final embedding and the Glasgow Coma Scale (GCS), which assesses the level of consciousness after a brain injury. This connection aligns well with common clinical knowledge, as the GCS score is critical in evaluating the severity of brain injuries and predicting functional independence after ICH onset~\cite{rost2008prediction}.
Attention Head 4 predominantly captured the relationship between the final embedding and the NIH Stroke Scale (NIHSS) score, which indicates the severity of stroke. The strong focus on the NIHSS score underscores its importance in predicting the prognosis of functional independence after ICH onset. Both the GCS and NIHSS are well-known predictors of clinical outcomes after ICH~\cite{hemphill2001score}. Attention matrices corresponding to the two most prominent self-attention heads are visualized in Figure~\ref{fig:sa_head1} and Figure~\ref{fig:sa_head2}.

\textbf{Important covariates for each treatment}. We also explored cross-attention between the covariate and treatment encoders to identify which covariates are important for outcome prediction for each treatment. Specifically, we analyzed the top covariates that had high attention weights for the [CLS] token for each treatment. For ATACH-II, the key covariates identified, in decreasing order of importance, were initial systolic and diastolic blood pressure (SBP and DBP), GCS score, Hispanic ethnicity, and White race. High SBP is a critical factor in intracerebral hemorrhage (ICH) outcomes, and the ATACH II trial specifically targeted SBP management to prevent hematoma expansion and improve clinical outcomes. While DBP is less frequently highlighted compared to SBP, it still contributes to overall cardiovascular stress and can influence outcomes. The GCS score is a well-established predictor of ICH outcomes. For MISTIE-III, the important covariates identified were Platelet Count, Sodium, Potassium, CO2, BUN, APTT, WBC Count, and Chloride level. These variables are crucial for patient recovery after surgical interventions like those in MISTIE-III. For instance, platelet count is essential for assessing the risk of bleeding and the body's ability to form clots; low platelet counts can increase the risk of hemorrhage during and after surgery~\cite{ziai2003platelet}. Both these analyses confirm that the patient representations learned by the covariate encoder, and the treatment effect captured by the cross-attention encoder indeed capture significant treatment and disease-level information. A visualization of the cross-attention activations is included in Figure~\ref{fig:ca_head}.



%% file: sections/conclusion.tex
This paper introduces a novel framework for ITE estimation that demonstrates efficacy in both centralized and distributed data settings, while accommodating heterogeneity in covariates, treatments, and outcome spaces. Notably, the framework extends its utility to zero-shot learning scenarios, allowing for the estimation of treatment effects for new and unseen interventions by leveraging supplementary treatment information. The findings of this study underscore two key insights: i) collaborative learning through federated mechanisms across diverse institutions enhances the quality of estimators, and ii) effective utilization of supplementary treatment information can facilitate accurate estimation of novel treatment effects. Although this paper primarily addresses healthcare applications, the proposed framework is applicable to other domains requiring ITE estimation. While the federated mechanism does not entail explicit data sharing, the method may encounter limitations in settings where sharing model parameters outside the client site, even with a trusted server, poses privacy risks. Therefore, a privacy-preserving version of the method will be considered for future work.

%% file: sections/appendix.tex
\begin{center}
\textbf{\Large Supplement for \enquote{Leveraging Data from Disparate Sources for ITE Estimation in Federated Learning Settings}}
\end{center}

In this supplementary material, we first provide additional details about the datasets used in the experiments, followed by the precise definitions of the evaluation metrics. We then include additional experimental results related to the local training of the models and visualizations associated with the interpretability analysis presented in the subsection~\ref{subsec:interpret}.

\section{Datasets}\label{app:datasets}
We first discuss the additional details of the semi-synthetic datasets. The IHDP dataset involves a binary treatment scenario with real covariates and simulated outcomes. In this dataset, covariates are obtained from both the mother and the child, with explicit child care or specialist home visits treated as interventions. The outcomes are the future cognitive test scores of the children. This dataset includes 25 covariates and approximately $\sim$743 datapoints.
The ACIC-2016 dataset, originally introduced for a competition, was sourced from the Collaborative Perinatal Project. It includes 55 covariates and $\sim$4,802 datapoints. The Twins dataset contains 39 covariates and $\sim$11,400 datapoints. It represents a real-world collection of twin births in the US between 1989 and 1991 and includes covariates related to the parents, pregnancy, and birth. In this dataset, birth weight is considered the treatment variable, while one-year mortality serves as the outcome, making it a binary treatment problem with binary outcomes. All three datasets were pre-processed using the processing method employed in CATENets~\cite{catenets}. 

For the first real-world dataset, we collect data from three randomized clinical trials for intracerebral hemorrhage (ICH) therapy development: ATACH2 (NCT01176565), MISTIE3 (NCT01827046), and ERICH (NCT01202864). Each hospital location provides patient-level pre-treatment measurements considered as covariates, such as demographics and clinical presentation of the ICH. A treatment variable indicates whether the patient is receiving active treatment or standard of care, and the outcomes are measured using the modified Rankin Score (mRS), which represents the patient's severity. Each of the three treatments is administered at a different hospital location, and all three trials include standard-of-care therapy. The second real-world dataset is obtained from the clinical trials for the Alzheimer's disease. The covariates include the pre-treatment measurements of the patients, and the outcome is the total ADAS-cognitive score of the patient. These real-world datasets are summarized in Table~\ref{tab:dataset_desc}.

\begin{table*}[!ht]
    \centering
    \caption{Dataset statistics and brief description for the real-world datasets used for experimental evaluation.}
    \label{tab:dataset_desc}
    \begin{center}
    \begin{small}
    \hspace*{-0.8cm}
    \addtolength{\tabcolsep}{-0.4em}
    \begin{tabular}{@{}lccccc@{}}
        \toprule
            &  \# subjects & \# covariates  &  \# treatments & \# sites  & Description \\[0.5ex]
         \midrule
         ICH data & 3279 & 45 & 3 & 3 & \multirowcell{2}[0pt][c]{Clinical trial data from 3 trials for therapy \\ development  of intracerebral hemorrhage, prepared in~\cite{ich_dataset}.}\\[0.5ex]
            & & & & & \\[0.5ex]
         CPAD data&  9406 & 144 & 7 & 19 & \multirowcell{2}[0pt][c]{Clinical trial data for Alzheimer's disease \\ therapy development from 38 Phase II/III trials~\cite{cpad_dataset}.}\\[0.5ex]
            & & & & & \\
         \bottomrule
    \end{tabular}
    \end{small}
    \end{center}
\end{table*}

\subsection{Zero-shot Inference}
As discussed in the Experiments section under subsection~\ref{subsec:zero_shot}, we use additional information to test the approach in the zero-shot setting. Since treatment names are available only for the ICH dataset, we can obtain the necessary additional information and test our method on this dataset. The information that we consider involves textual descriptions of the treatments in natural language, and is sourced from treatment descriptions provided on the website clinicaltrials.gov under ARM details for the treatment. On average, these descriptions contain 55 words. To give a clearer understanding of what these descriptions entail, we highlight excerpts from them and present them in Table~\ref{tab:treatment_desc} below.

\begin{table*}[!ht]
    \centering
    \caption{Excerpts of the descriptions obtained from clinicaltrials.gov for ATACH2, MISTIE and Placebo treatments used in zero-shot inference experiments.}
    \label{tab:treatment_desc}
    \begin{center}
    \begin{small}
    \hspace*{-0.8cm}
    \addtolength{\tabcolsep}{-0.4em}
    \begin{tabular}{@{}clc@{}}
        \toprule
         Treatment Name & \multicolumn{1}{c}{Description Snippets} \\
         \midrule
          ATACH-II & \multirowcell{3}[0pt][c]{Antihypertensive Treatment of Acute Cerebral Hemorrhage II uses early intensive blood pressure lowering. Intravenous  \\
          nicardipine hydrochloride will be used as necessary (pro re nata or "PRN") as the primary agent in lowering SBP. The goal for the \\ 
          intensive BP reduction group will be to reduce and maintain SBP < 140 mmHg for 24 hours from randomization.}\\[0.5ex]
            &  \\[0.5ex]
          &  \\[0.5ex]
         MISTIE-III & \multirowcell{3}[0pt][c]{Subjects randomized to the Minimally Invasive Surgery (MIS) plus rt-PA management arm will undergo minimally \\ invasive surgery followed by up to 9 doses of 1.0 mg of rt-PA (Activase/Alteplase/CathFlo) for intracerebral hemorrhage \\ clot resolution.}\\[0.5ex]
           &  \\[0.5ex]
             &  \\[0.5ex]
        Placebo & \multirowcell{3}[0pt][c]{Subjects randomized to medical management will receive the standard medical therapies for the treatment \\ of intracerebral hemorrhage, which includes ICU care only and no planned surgical intervention.}\\[0.5ex]
           &  \\[0.5ex]
             &  \\[0.5ex]
         \bottomrule
    \end{tabular}
    \end{small}
    \end{center}
\end{table*}

\section{Metrics}\label{app:metrics}
In this section, we precisely define the metrics used for evaluating our methods against the baselines. Since, we consider a multi-treatment setting, we show the generalization of the metric definitions for any treatment denoted by $j$ and calculated over $n$ subjects (datapoints). Firstly, we show the metrics applicable when true treatment effects can be calculated by using the factual and counterfactual outcomes available. These metrics are used for evaluation on the semi-synthetic datasets.

$$ \text{ATE}_{\epsilon}(j) = \frac{1}{n} \sum_{i=1}^{n} (\mu_j(x_i) - \mu_0(x_i)) - \frac{1}{n}\sum_{i=1}^{n} (\hat{\mu}_j(x_i) - \hat{\mu}_0(x_i)).$$
$$ \text{PEHE}(j) = \frac{1}{n}\sum_{i=1}^{n} \bigg( (\mu_j(x_i) - \mu_0(x_i)) - (\hat{\mu}_j(x_i) - \hat{\mu}_0(x_i)) \bigg)^2.$$

Then, we introduce the metrics applicable in real-world settings, where only factual outcomes are available and the true treatment effect cannot be computed.
$$\text{RMSE-F} = \sqrt{ \frac{1}{n} \sum_{i=1}^{n} \sum_{j=1}^{K} \mathbbm{1}(\T_i=j) (\Y_i(j) - \hat{\mu}_j(x_i))^2 }.$$
$$\text{ATT}(j) = \dfrac{1}{|T_j|} \sum_{i=1}^{n}\mathbbm{1}(\T_i=j) \Y_i(j) - \dfrac{1}{|T_0|} \sum_{i=1}^{n}\mathbbm{1}(\T_i=0) \Y_i(0)$$
$$\text{ATT}_{\epsilon}(j) = | \text{ATT}(j ) - \dfrac{1}{|T_j|} \sum_{i=1}^{n}\mathbbm{1}(\T_i=j) (\hat{\mu}_j(x_i) - \hat{\mu}_0(x_i)) |$$
where $|T_j|$ and $|T_0|$ represents the count of patients receiving treatment $j$ and treatment $0$ respectively.

\section{Additional Experiments}\label{app:experiments}
This section provides additional experimental results to complement those presented in Table~\ref{table:performance_syn}. These results are obtained in the training protocol of local model training. This setting is highlighted because when the methods do not suggest a way of collaboration across multiple sites, each site can only train its own local model for ITE estimation using its respective dataset. For comparison, we also show the result obtained if our method FedTransTEE under federated learning protocol is used instead.

\begin{table*}[!ht]
\centering
\caption{The results in the table show performance comparison between our method and the related ITE prediction methods on the held-out test dataset for the local setting on semi-synthetic datasets.}
\label{table:performance_syn_local}
\setlength\tabcolsep{3pt}
\begin{center}
\begin{small}
\scalebox{0.9}{
\hspace*{-1.5cm}
\begin{tabular}{@{}lccccccccc@{}}
\toprule
 \multirow{2}{*}[-0.5ex]{Method} & \multicolumn{3}{c}{IHDP} & \multicolumn{3}{c}{ACIC-16} & 
 \multicolumn{3}{c}{Twins} \\
   \cmidrule(lr){2-4} \cmidrule(lr){5-7} \cmidrule(lr){8-10}
        & (PEHE) & (ATE$_\epsilon$) & (RMSE-F) & (PEHE) & (ATE$_\epsilon$) & (RMSE-F) & (PEHE) & (ATE$_\epsilon$) & (RMSE-F) \\[0.5ex]
 \midrule
 (Local) & & & & & & & & & \\
 S-Learner$_{l}$ & 1.11 $\pm$ 0.02 & 3.3 $\pm$ 0.06 &  1.83 $\pm$ 0.09 &  2.2$\pm$0.01 &  2.9$\pm$0.02  &  2.3$\pm$0.1 & \textbf{0.32$\pm$0.01} & 0.02$\pm$0.001  &   0.10$\pm$0.02 \\ [0.3ex]
 T-Learner$_{l}$ & 1.5 $\pm$ 0.06 & 3.58 $\pm$ 0.13 &  \textbf{1.55 $\pm$ 0.1} &   2.8$\pm$0.02  &  3.3$\pm$0.02  &  1.4$\pm$0.01  &  0.35$\pm$0.001 & 0.04$\pm$0.001 &   0.11$\pm$0.001 \\ [0.3ex]
 TARNet$_{l}$ & 1.22 $\pm$ 0.02 & 3.7 $\pm$ 0.06 & 1.59 $\pm$ 0.1 &  2.77$\pm$0.01  &  3.3$\pm$0.02  &  3.4$\pm$0.02  &  0.35$\pm$0.001  &  0.03$\pm$0.01  &   0.11$\pm$0.005 \\ [0.3ex]
 FlexTENet$_{l}$ & 1.22 $\pm$ 0.2 & 3.7 $\pm$ 0.06 & 1.6 $\pm$ 0.07 &   2.6$\pm$0.02  &  3.34$\pm$0.01  &  1.7$\pm$0.01  &  0.33$\pm$0.001  &  0.02$\pm$0.001  &   \textbf{0.09$\pm$0.002} \\ [0.3ex]
 HyperITE$_{l}$ (S-Learner) & \textbf{1.02 $\pm$ 0.03} & 3.59 $\pm$ 0.1 & 1.87 $\pm$ 0.23 &  2.2$\pm$0.01  &   3.1$\pm$0.02  &  2.84$\pm$0.05  &  \textbf{0.32$\pm$0.001}  &  0.02$\pm$0.001  &  0.11$\pm$0.004 \\ [0.3ex]
 HyperITE$_{l}$ (TARNet) & 1.06 $\pm$ 0.02 & 3.8 $\pm$ 0.06 &  1.6 $\pm$ 0.07 &  2.7 $\pm$ 0.02  &  3.1 $\pm$ 0.03  &  3.4 $\pm$ 0.01  & 0.33 $\pm$ 0.001 & 0.03 $\pm$ 0.002  &  0.10 $\pm$ 0.002 \\ [0.3ex]
  \midrule
 FedTransTEE (Ours) & \textbf{1.02 $\pm$ 0.01} & \textbf{0.26 $\pm$ 0.06} & 1.77 $\pm$ 0.5 &  \textbf{0.78 $\pm$ 0.1} &  \textbf{0.326 $\pm$ 0.02} & \textbf{0.728 $\pm$ 0.01} & \textbf{0.32 $\pm$ 0.01} & \textbf{0.01 $\pm$ 0.002} &  \textbf{0.09 $\pm$ 0.01} \\ [0.3ex]
\bottomrule
\end{tabular}
}
\end{small}
\end{center}    
\end{table*}

\subsection{Attention Visualizations}\label{app:attention_visual}
We visualize and analyze the cross-attention and self-attention heads' activations obtained during the learning process of FedTransTEE. This allows us to identify the covariates that are crucial in predicting outcomes for different treatments and to explore the relationships between these covariates. Such examination underscores both the explainability and interpretability of our approach. While detailed analysis is provided in subsection~\ref{subsec:interpret}, the visualizations are included below.

\begin{figure}
\centering
    \centering
    \includegraphics[scale=0.35]{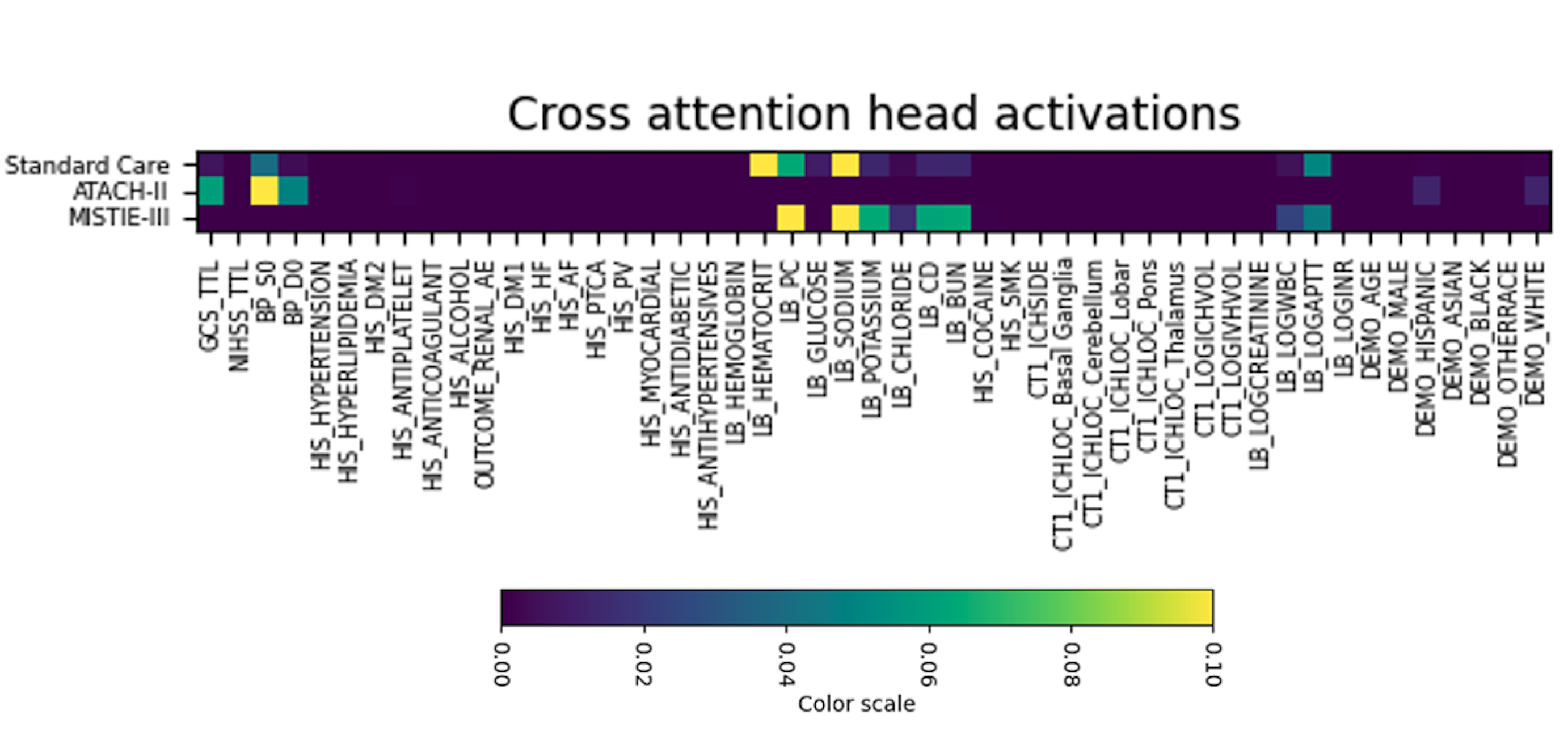}
    \caption{Visualization of the activations of cross-attention head obtained while learning on the ICH dataset.}
    \label{fig:ca_head}
\end{figure}

\begin{figure}
\centering
    \centering
    \includegraphics[scale=0.65]{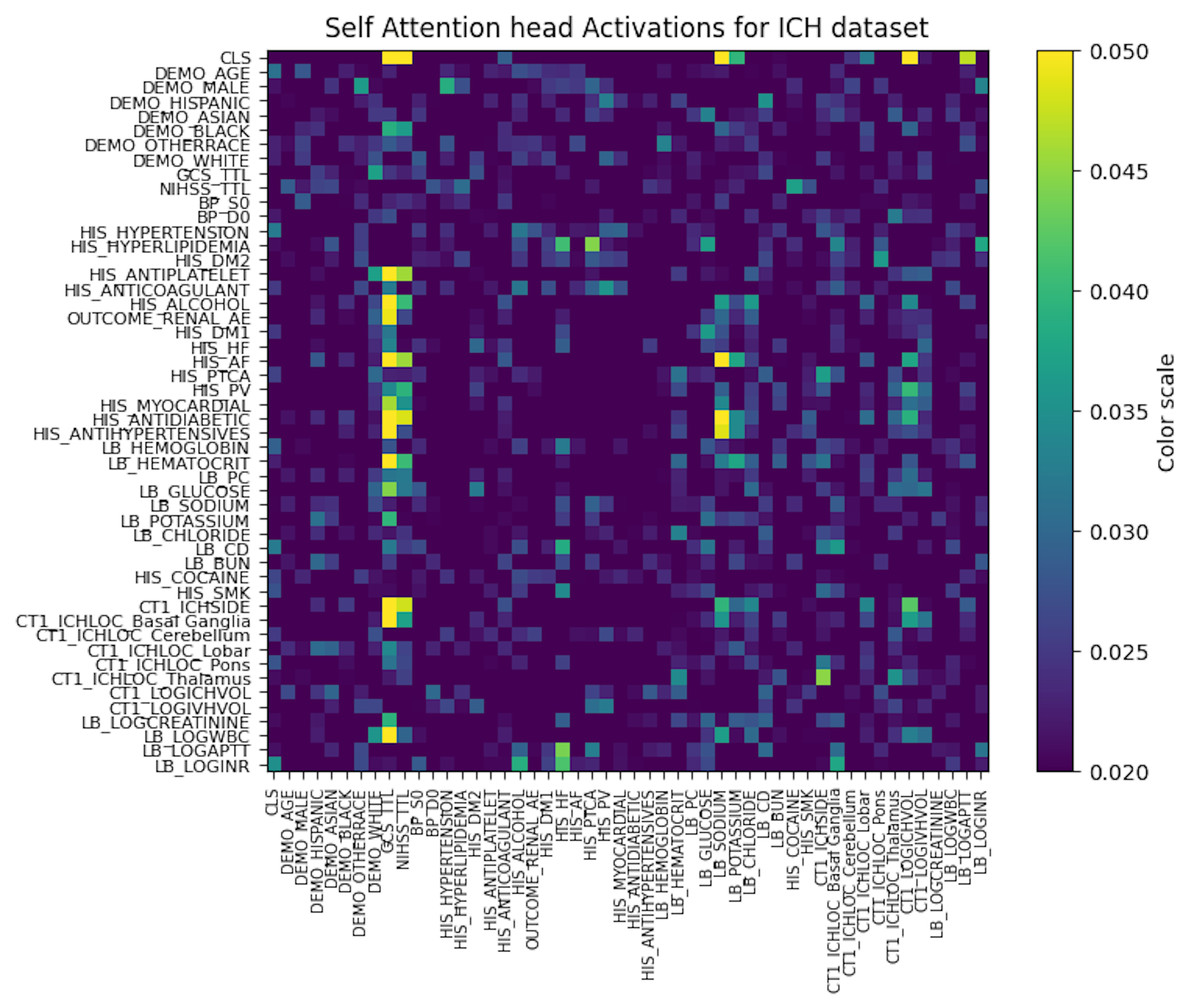}
    \caption{Visualization of the activations of first self-attention head of the covariate encoder obtained while learning on the ICH dataset.}
    \label{fig:sa_head1}
\end{figure}

\begin{figure}
\centering
    \centering
    \includegraphics[scale=0.65]{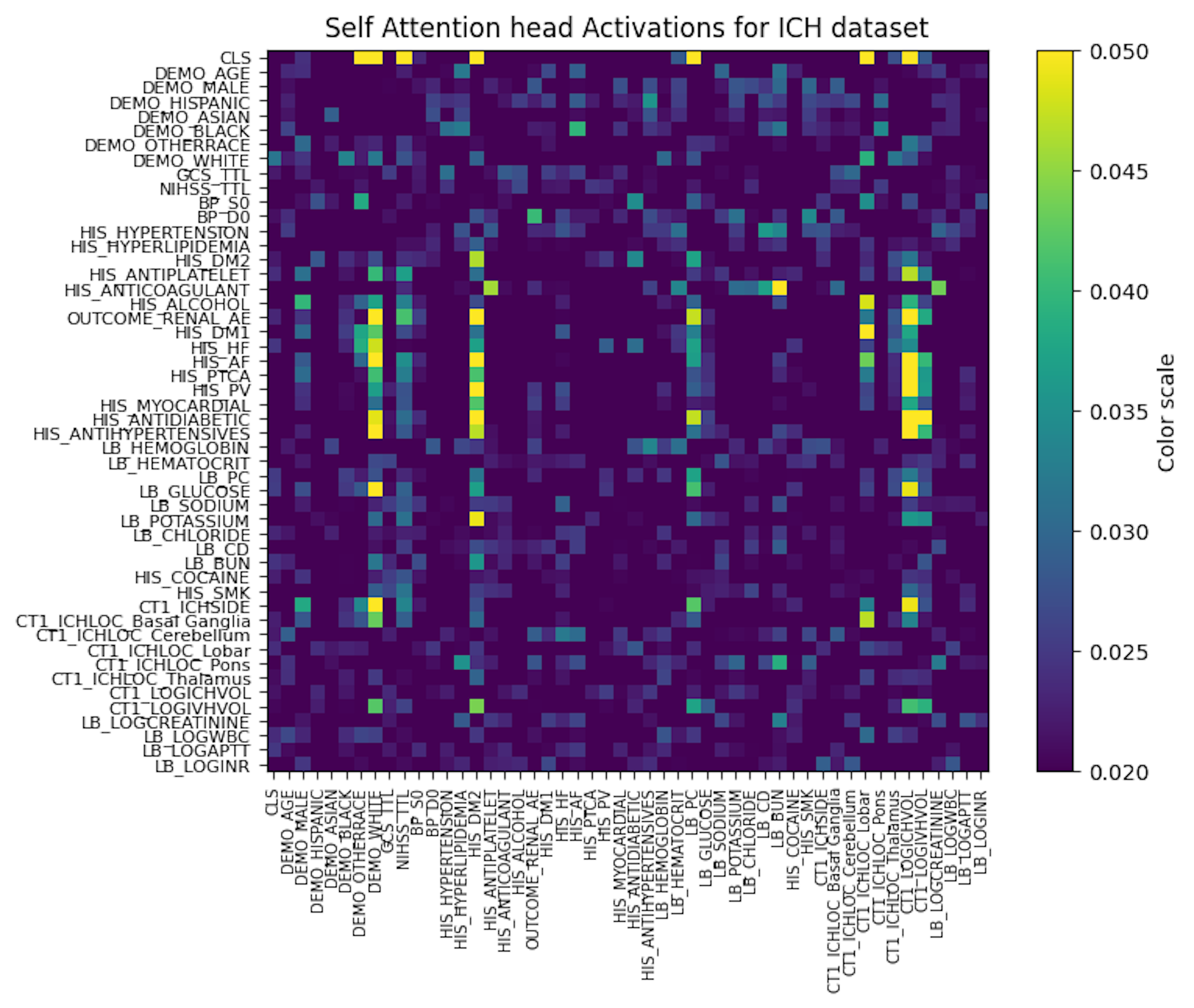}
    \caption{Visualization of the activations of second self-attention head of the covariate encoder obtained while learning on the ICH dataset.}
    \label{fig:sa_head2}
\end{figure}